# Rerepresenting and Restructuring Domain Theories: A Constructive Induction Approach

**Steven K. Donoho**                                               DONOHO@CS.UIUC.EDU
**Larry A. Rendell**                                               RENDELL@CS.UIUC.EDU
*Department of Computer Science, Univeristy of Illinois*
*405 N. Mathews Ave., Urbana, IL 61801 USA*

## Abstract

Theory revision integrates inductive learning and background knowledge by combining training examples with a coarse domain theory to produce a more accurate theory. There are two challenges that theory revision and other theory-guided systems face. First, a representation language appropriate for the initial theory may be inappropriate for an improved theory. While the original representation may concisely express the initial theory, a more accurate theory forced to use that same representation may be bulky, cumbersome, and difficult to reach. Second, a theory structure suitable for a coarse domain theory may be insufficient for a fine-tuned theory. Systems that produce only small, local changes to a theory have limited value for accomplishing complex structural alterations that may be required.

Consequently, advanced theory-guided learning systems require *flexible representation* and *flexible structure*. An analysis of various theory revision systems and theory-guided learning systems reveals specific strengths and weaknesses in terms of these two desired properties. Designed to capture the underlying qualities of each system, a new system uses theory-guided constructive induction. Experiments in three domains show improvement over previous theory-guided systems. This leads to a study of the behavior, limitations, and potential of theory-guided constructive induction.

## 1. Introduction

Inductive learners normally use training examples, but they can also use background knowledge. Effectively integrating this knowledge into induction has been a widely studied research problem. Most work to date has been in the area of *theory revision* in which the knowledge given is a coarse, perhaps incomplete or incorrect, theory of the problem domain, and training examples are used to shape this initial theory into a refined, more accurate theory (Ourston & Mooney, 1990; Thompson, Langley, & Iba, 1991; Cohen, 1992; Pazzani & Kibler, 1992; Baffes & Mooney, 1993; Mooney, 1993). We develop a more flexible and more robust approach to the problem of learning from both data and theory knowledge by addressing the two following desirable qualities:

- *Flexible Representation.* A theory-guided system should utilize the knowledge contained in the initial domain theory without having to adhere closely to the initial theory's representation language.

- *Flexible Structure.* A theory-guided system should not be unnecessarily restricted by the structure of the initial domain theory.

Before giving more precise definitions of our terms, we motivate our work intuitively.

## 1.1 Intuitive Motivation

The first desirable quality, flexibility of representation, arises because the theory representation most appropriate for describing the coarse, initial domain theory may be inadequate for the final, revised theory. While the initial domain theory may be compact and concise in one representation, an accurate theory may be quite bulky and cumbersome in that representation. Furthermore, the representation that is best for expressing the initial theory may not be the best for *carrying out* refinements. A helpful refinement step may be clumsy to make in the initial representation yet be carried out quite simply in another representation.

As a simple example, a coarse domain theory may be expressed as the logical conjunction of N conditions that should be met. The most accurate theory, though, is one in which any M of these N conditions holds. Expressing this more accurate theory in the DNF representation used to describe the initial theory would be cumbersome and unwieldy (Murphy & Pazzani, 1991). Furthermore, arriving at the final theory using the refinement operators most suitable for DNF (drop-condition, add-condition, modify-condition) would be a cumbersome task. But when an M-of-N representation is adopted, the refinement simply involves empirically finding the appropriate M, and the final theory can be expressed concisely (Baffes & Mooney, 1993).

Similarly, the second desirable quality, flexibility of structure, arises because the theory structure that was suitable for a coarse domain theory may be insufficient for a fine-tuned theory. In order to achieve the desired accuracy, a restructuring of the initial theory may be necessary. Many theory revision systems act by making a series of local changes, but this can lead to behavior at two extremes. The first extreme is to rigidly retain the backbone structure of the initial domain theory, only allowing small, local changes. Figure 1 illustrates this situation. Minor revisions have been made – conditions have been added, dropped, and modified – but the refined theory is trapped by the backbone structure of the initial theory. When only local changes are needed, these techniques have proven useful (Ourston & Mooney, 1990), but often more is required. When more *is* required, these systems often move to the other extreme; they drop entire rules and groups of rules and then build entire new rules and groups of rules from scratch to replace them. Thus they restructure, but they forfeit valuable knowledge in the process. An ideal theory revision system would glean knowledge from theory substructures that cannot be fixed with small, local changes and use this in a restructured theory.

As an intuitive illustration, consider a piece of software that "almost works." Sometimes it can be made useful through only a few local operations: fixing a couple of bugs, adding a needed subroutine, and so on. In other cases, though, a piece of software that "almost works" is in fact *far* from *full* working order. It may need to be redesigned and restructured. A mistake at one extreme is to try to fix a program like this by making a series of patches in the original code. A mistake at the other extreme is to discard the original program without learning anything from it and start from scratch. The best approach would be to examine the original program to see what can be learned from its design and to use this knowledge in the redesign. Likewise, attempting to improve a coarse domain theory through a series of local changes may yield little improvement because the theory is trapped by its initial structure. This does not render the original domain theory useless; careful analysis of the initial domain theory can give valuable guidance for the design of the best final theory. This is illustrated in Figure 2 where many substructures have been taken from the initial theory and adapted for use in the refined theory. Information from the initial theory has been used, but the structure of the revised theory is not restricted by the structure of the initial theory.

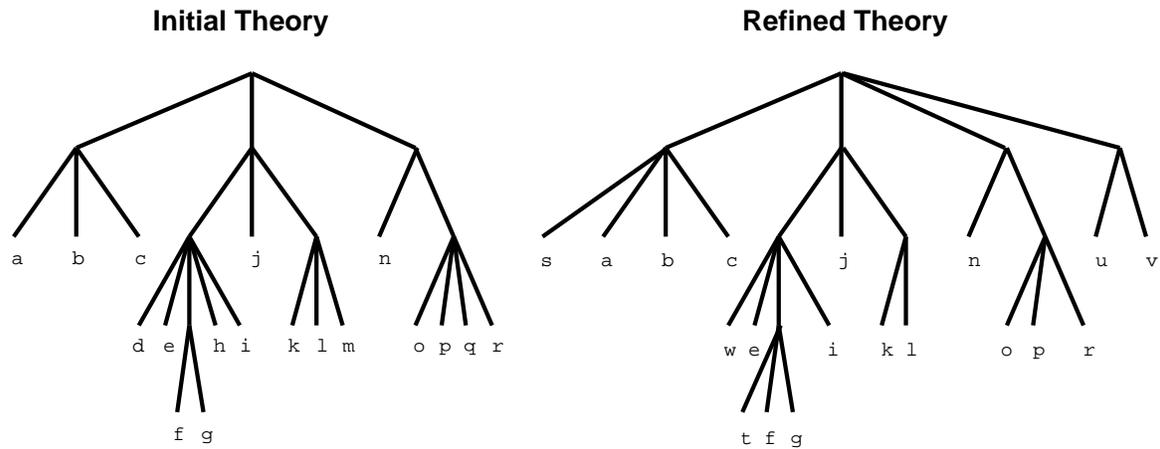

Figure 1: Typical theory revision allows only limited structural flexibility. Although conditions have been added, dropped, and modified, the revised theory is much constrained by the structure of the initial theory.

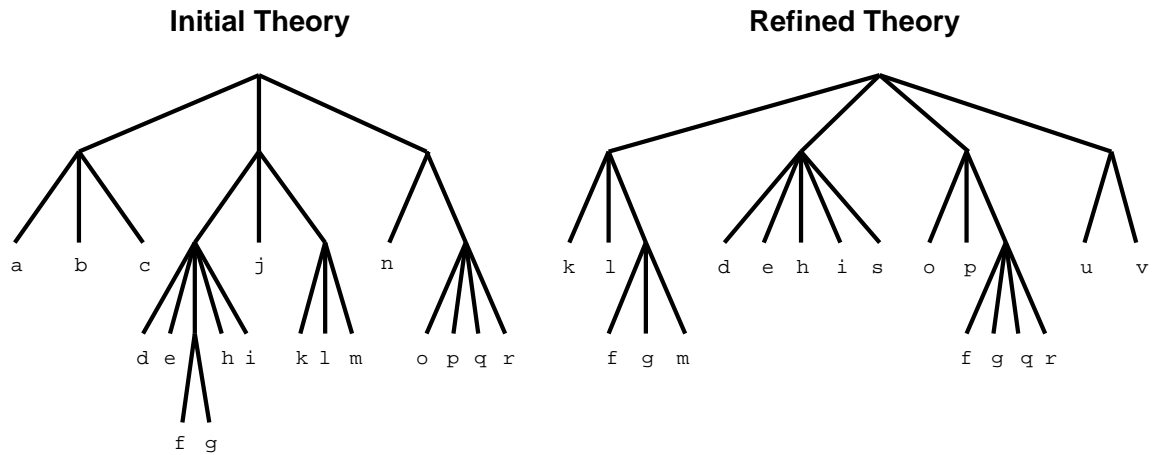

Figure 2: More flexible structural modification. The revised theory has taken many substructures from the initial theory and adapted and recombined them for its use, but the structure of the revised theory is not restricted by the structure of the initial theory.

Donoho & Rendell## 1.2 Terminology

In this paper, all *training data* consist of *examples* which are classified vectors of feature/value pairs. We assume that an *initial theory* is a set of conditions combined using the operators AND, OR, and NOT and indicating one or more classes. While it is unreasonable to believe that all theories will always be of this form, it covers much existing theory revision research.

Our work is intended as an informal exploration of flexible representation and flexible structure. *Flexible representation* means allowing the theory to be revised using a representation language other than that of the initial theory. An example of flexible representation is the introduction of a new operator for combining features — an operator not used in the initial theory. In Section 1.1 the example was given of introducing the M-of-N operator to represent a theory originally expressed in DNF. *Flexible structure* means not limiting revision of the theory to a series of small, incremental modifications. An example of this is breaking the theory down into its components and using them as building blocks in the construction of a new theory.

*Constructive induction* is a process whereby the training examples are redescribed using a new set of features. These new features are combinations of the original features. Bias or knowledge may be used in the construction of the new features. A subtle point is that when we speak of flexible representation, we are referring *only* to the representation of the domain theory, *not* the training data. Although the phrase "change of representation" is often applied to constructive induction, this refers to a change of the data. In our paper, the term flexible representation is reserved for a change of theory representation. Thus a system can be performing constructive induction (changing the feature language of the data) without exhibiting flexible representation (changing the representation of the theory).

## 1.3 Overview

Theory revision and constructive induction embody complementary aspects of the machine learning research community's ultimate goals. Theory revision uses data to improve a theory; constructive induction can use a theory to improve data to facilitate learning. In this paper we present a theory-guided constructive induction approach which addresses the two desirable qualities discussed in Section 1.1. The initial theory is analyzed, and new features are constructed based on the components of the theory. The constructed features need not be expressed in the same representational language as the initial theory and can be refined to better match the training examples. Finally, a standard inductive learning algorithm, C4.5 (Quinlan, 1993), is applied to the redescribed examples.

We begin by analyzing how landmark theory revision and learning systems have exhibited flexibility in handling a domain theory and what part this has played in their performance. From this analysis, we extract guidelines for system design and apply them to the design of our own limited system. In an effort to integrate learning from theory and data, we borrow heavily from the theory revision, multistrategy learning, and constructive induction communities, but our guidelines for system design fall closest to classical constructive induction methods. The central focus of this paper is not the presentation of "another new system" but rather a study of flexible representation and structure, their manifestation in previous work, and their guidance for future design.

414

Section 2 gives the context of our work by analyzing previous research and its influence on our work. Section 3 explores the Promoter Recognition domain and demonstrates how related theory revision systems behave in this domain. In Section 4, guidelines for theory-guided constructive induction are presented. These guidelines are a synthesis of the positive aspects of related research, and they address the two desirable qualities, flexibility of representation and flexibility of structure. Section 4 also presents a specific theory-guided constructive induction algorithm which is an instantiation of the guidelines set forth earlier in that section. Results of experiments in three domains are given in Section 5 followed by a discussion of the strengths of theory-guided constructive induction in Section 6. Section 7 presents an experimental analysis of the limits of applicability of our simple algorithm followed by a discussion of limitations and future directions of our work in Section 8.

## 2. Context and Related Work

Although our work bears some resemblance in form and objective to many papers in constructive induction (Michalski, 1983; Fu & Buchanan, 1985; Utgoff, 1986; Schlimmer, 1987; Drastal & Raatz, 1989; Matheus & Rendell, 1989; Pagallo & Haussler, 1990; Ragavan & Rendell, 1993; Hirsh & Noordewier, 1994), theory revision (Ourston & Mooney, 1990; Feldman, Serge, & Koppel, 1991; Thompson et al., 1991; Cohen, 1992; Pazzani & Kibler, 1992; Baffes & Mooney, 1993), and multistrategy approaches (Flann & Dietterich, 1989; Towell, Shavlik, & Noordeweir, 1990; Dzerisko & Lavrac, 1991; Bloedorn, Michalski, & Wnek, 1993; Clark & Matwin, 1993; Towell & Shavlik, 1994), we focus only upon a handful of these systems, those that have significant, underlying similarities to our work. In this section we analyze Miro, Either, Focl, Labyrinth$_K$, Kbann, Neither-MofN, and Grendel to discuss their related underlying contributions in relationship to our perspective.

### 2.1 Miro

Miro (Drastal & Raatz, 1989) is a seminal work in knowledge-guided constructive induction. It takes knowledge about how low-level features interact and uses this knowledge to construct high-level features for its training examples. A standard learning algorithm is then run on these examples described using the new features. The domain theory is used to shift the bias of the induction problem (Utgoff, 1986). Empirical results showed that describing the examples in these high-level, abstract terms improved learning accuracy.

The Miro approach provides a means of utilizing knowledge in a domain theory without being restricted by the structure of that theory. Substructures of the domain theory can be used to construct high-level features that a standard induction algorithm will arrange into a concept. Some constructed features will be used as they are, others will be ignored, others will be combined with low-level features, and still others may be used differently in multiple contexts. The end result is that knowledge from the domain theory is utilized, but the structure of the final theory is not restricted by the structure of the initial theory. Miro provides flexible structure.

Another benefit is that Miro-like techniques can be applied even when only a partial domain theory exists, i.e., a domain theory that only specifies high-level features but does not link them together or a domain theory that specifies some high-level features but not others. One of Miro's shortcomings is that it provided no means of making minor changes in the domain theory but rather constructed the features exactly as the domain theory specified. Also the representation of MIRO's constructed features was primitive — either an example met the conditions of a high-level feature or did not. An example of MIRO's behavior is given in Section 3.2.

## 2.2 Either, Focl, and Labyrinth$_K$

The EITHER (Ourston & Mooney, 1990), LABYRINTH$_K$ (Thompson et al., 1991), and FOCL (Pazzani & Kibler, 1992) systems represent a broad spectrum of theory revision work. They make steps toward effective integration of background knowledge and inductive learning. Although these systems have many superficial differences with regard to supervised/unsupervised learning, concept description language, etc., they share the underlying principle of incrementally revising an initial domain theory through a series of local changes.

We will discuss EITHER as a representative of this class of systems. EITHER's theory revision operators include: removing unwanted conditions from a rule, adding needed conditions to a rule, removing rules, and adding totally new rules. EITHER first classifies its training examples according to the current theory. If any are misclassified, it seeks to repair the theory by applying a theory revision operator that will result in the correct classification of some previously misclassified examples without losing any of the correct examples. Thus a series of local changes are made that allow for an improvement of accuracy on the training set without losing any of the examples previously classified correctly.

EITHER-type methods provide simple yet powerful tools for repairing many important and common faults in domain theories, but they fail to meet the qualities of flexible representation and flexible structure. Because the theory revision operators make small, local modifications in the existing domain theory, the final theory is constrained to be similar in structure to the initial theory. When an accurate theory is significantly different in structure from the initial theory, these systems are forced to one of the two extremes discussed in Section 1. The first extreme is to become trapped at a local maximum similar to the initial theory unable to reach the global maximum because only local changes can be made. The other extreme is to drop entire rules and groups of rules and replace them with new rules built from scratch thus forfeiting the knowledge contained in the domain theory.

Also, EITHER carries out all theory revision steps in the representation of the initial theory. Consequently, the representation of the final theory is the same as that of the initial theory. Another representation may be more appropriate for the revised theory than the one in which the initial theory comes, but facilities are not provided to accommodate this. An advanced theory revision system would combine the locally acting strengths of EITHER-type systems with flexibility of structure and flexibility of representation. An example of EITHER's behavior is given in Section 3.3.

## 2.3 Kbann and Neither-MofN

The KBANN system (Towell et al., 1990; Towell & Shavlik, 1994) makes unique contributions to theory revision work. KBANN takes an initial domain theory described symbolically in logic and creates a neural network whose structure and initial weights encode this theory. Backpropagation (Rumelhart, Hinton, & McClelland, 1986) is then applied as a refinement tool for fine-tuning the network weights. KBANN has been empirically shown to give significant improvement over many theory revision systems for the widely-used Promoter Recognition domain. Although our work is different in implementation from KBANN, our abstract ideologies are similar.

One of KBANN's important contributions is that it takes a domain theory in one representation (propositional logic) and translates it into a less restricting representation (neural network). While logic is an appropriate representation for the initial domain theory for the promoter problem, the neural network representation is more convenient both for refining this theory and for expressing the best revised theory. This change of representation is KBANN's real source of power. Much attention has been given to the fact that KBANN combines symbolic knowledge with a subsymbolic learner, but this combination can be viewed more generally as a means of implementing the important change of representation. It may be the change of representation that gives KBANN its power, not necessarily its specific symbolic/subsymbolic implementation. Thus the KBANN system embodies the higher-level principle of allowing refinement to occur in an appropriate representation.

If an alternative representation is KBANN's source of power, the question must be raised as to whether the actual KBANN implementation is always the best means of achieving this goal. The neural network representation may be more expressive than is required. Accordingly, backpropagation often has more refinement power than is needed. Thus KBANN may carry excess baggage in translating into the neural net representation, performing expensive backpropagation, and extracting symbolic rules from the refined network. Although the full extent of KBANN's power may be needed for some problems, many important problems may be solvable by applying KBANN's principles at the symbolic level using less expensive tools.

NEITHER-MOFN (Baffes & Mooney, 1993), a descendant of EITHER, is a second example of a system that allows a theory to be revised in a representation other than that of the initial theory. The domain theory input into NEITHER-MOFN is expressed in propositional logic as an AND/OR tree. NEITHER-MOFN interprets the theory less rigidly — an AND rule is true any time any M of its N conditions are true. Initially M is set equal to N (all conditions must be true for the rule to be true), and one theory refinement operator is to lower M for a particular rule. The end result is that examples that are a close enough partial match to the initial theory are accepted. NEITHER-MOFN, since it is built upon the EITHER framework, also includes EITHER-like theory revision operators: add-condition, drop-condition, etc.

Thus NEITHER-MOFN allows revision to take place in a representation appropriate for revision and appropriate for concisely expressing the best refined theory. NEITHER-MOFN has achieved results comparable to KBANN on the Promoter Recognition domain, which suggests that it is the change of representation which these two systems share that give them their power rather than any particular implementation. NEITHER-MOFN also demonstrates that a small amount of representational flexibility is sometimes enough. The M-of-N representation it employs is not as big a change from the original representation as the neural net representation which KBANN employs yet it achieves similar results and arrives at them much more quickly than KBANN (Baffes & Mooney, 1993).

A shortcoming of NEITHER-MOFN is that since it acts by making local changes in an initial theory, it can still become trapped by the structure of the initial theory. An advanced theory revision system would incorporate NEITHER-MOFN's and KBANN's flexibility of representation and allow knowledge-guided theory restructuring. Examples of KBANN's and NEITHER-MOFN's behavior are given in Sections 3.4 and 3.5.

### 2.4 GRENDEL

Cohen (1992) analyzes a class of theory revision systems and draws some insightful conclusions. One is that "generality [in theory interpretation] comes at the expense of power." He draws this principle from the fact that a system such as EITHER or FOCL treats every domain theory the same and therefore must treat every domain theory in the most general way. He argues that rather than just applying the most general refinement strategy to every problem, a small set of refinement strategies should be available that are narrow enough to gain leverage yet not so narrow that they only apply to a single problem. Cohen presents GRENDEL, a toolbox of translators each of which transforms a domain theory into an explicit bias. Each translator interprets the domain theory in a different way, and the most appropriate interpretation is applied to a given problem.

We apply Cohen's principle to the representation of domain theories. If all domain theories are translated into the same representation, then the most general, adaptable representation has to be used in order to accommodate the most general case. This comes at the expense of higher computational costs and possibly lower accuracy due to overfit stemming from unbridled adaptability. The neural net representation into which KBANN translates domain theories allows 1) a measure of partial match to the domain theory 2) different parts of the domain theory to be weighted differently 3) conditions to be added to and dropped from the domain theory. All these options of adaptability are probably not *necessary* for most problems and may even be detrimental. These options in KBANN also require the computationally expensive backpropagation method.

The representation used in NEITHER-MOFN is not as adaptable as KBANN's — it does not allow individual parts of the domain theory to be weighted differently. NEITHER-MOFN runs more quickly than KBANN on small problems and probably matches or even surpasses KBANN's accuracy for many domains — domains for which fine-grained weighting is unfruitful or even detrimental. A toolbox of theory rerepresentation translators analogous to GRENDEL would allow a domain theory to be translated into a representation having the most appropriate forms of adaptability.

### 2.5 Outlook and Summary

In summary, we briefly reexamine *flexible representation* and *flexible structure*, the two desirable qualities set forth in Section 1. We consider how the various systems exemplify some subset of these desirable qualities.

- KBANN and NEITHER-MOFN both interpreted a theory more flexibly than its original representation allowed and revised the theory in this more adaptable representation. A final, refined theory often has many exceptions to the rule; it may tolerate partial matches and missing pieces of evidence; it may weight some evidence more heavily than other evidence. KBANN's and NEITHER-MOFN's new representation may not be the most concise, appropriate representation for the initial theory, but the new representation allows concise expression of an otherwise cumbersome final theory. These are cases of the principle of *flexible representation*.

- Standard induction programs have been quite successful at building concise theories with high predictive accuracy when the target concept *can* be concisely expressed using the original set of features. When it can't, constructive induction is a means of creating new features such that the target concept can be concisely expressed. MIRO uses constructive induction to take advantage of the strengths of both a domain theory and standard induction. Knowledge from the theory guides the construction of appropriate new features, and standard induction structures these into a concise description of the concept. Thus MIRO-like construction coupled with standard induction provides a ready and powerful means of flexibly restructuring the knowledge contained in an initial domain theory. This is a case of the principle of *flexible structure*.

In the following section we introduce the DNA Promoter Recognition domain in order to illustrate tangibly how some of the systems discussed above integrate knowledge and induction.

## 3. Demonstrations of Related Work

This section introduces the Promoter Recognition domain (Harley, Reynolds, & Noordewier, 1990) and briefly illustrates how a MIRO-like system, EITHER, KBANN, and NEITHER-MOFN behave in this domain. We implemented a MIRO-like system for the promoter domain; versions of EITHER and NEITHER-MOFN were available from Ray Mooney's group; KBANN's behavior is described by analyzing (Towell & Shavlik, 1994). We chose the promoter domain because it is a non-trivial, real-world problem which a number of theory revision researchers have used to test their work (Ourston & Mooney, 1990; Thompson et al., 1991; Wogulis, 1991; Cohen, 1992; Pazzani & Kibler, 1992; Baffes & Mooney, 1993; Towell & Shavlik, 1994). The promoter domain is one of three domains in which we evaluate our work, theory-guided constructive induction, in Section 5.

### 3.1 The Promoter Recognition Domain

A promoter sequence is a region of DNA that marks the beginning of a gene. Each example in the promoter recognition domain is a region of DNA classified either as a promoter or a non-promoter. As illustrated in Figure 3, examples consist of 57 features representing a sequence of 57 DNA nucleotides. Each feature can take on the values A,G,C, or T representing adenine, guanine, cytosine, and thymine at the corresponding DNA position. The features are labeled according to their position from p-50 to p+7 (there is no zero position). The notation "p-$N$" denotes the nucleotide that is $N$ positions upstream from the beginning of the gene. The goal is to predict whether a sequence is a promoter from its nucleotides. A total of 106 examples are available: 53 promoters and 53 non-promoters.

The promoter recognition problem comes with the initial domain theory shown in Figure 4 (quoted almost verbatim from Towell and Shavlik's entry in the UCI Machine Learning Repository). The theory states that promoter sequences must have two regions that make contact with a protein and must also have an acceptable conformation pattern. There are four possibilities for the contact region at *minus_35* (35 nucleotides upstream from the beginning of the gene). A match of any of these four possibilities will satisfy the *minus_35* contact condition, thus they are joined by disjunction. Similarly, there are four possibilities for the contact region at *minus_10* and four acceptable *conformation* patterns. Figure 5 gives a more pictorial presentation of portions of the theory. Of the 106 examples in the dataset, *none* matched the domain theory exactly, yielding an accuracy of 50%.

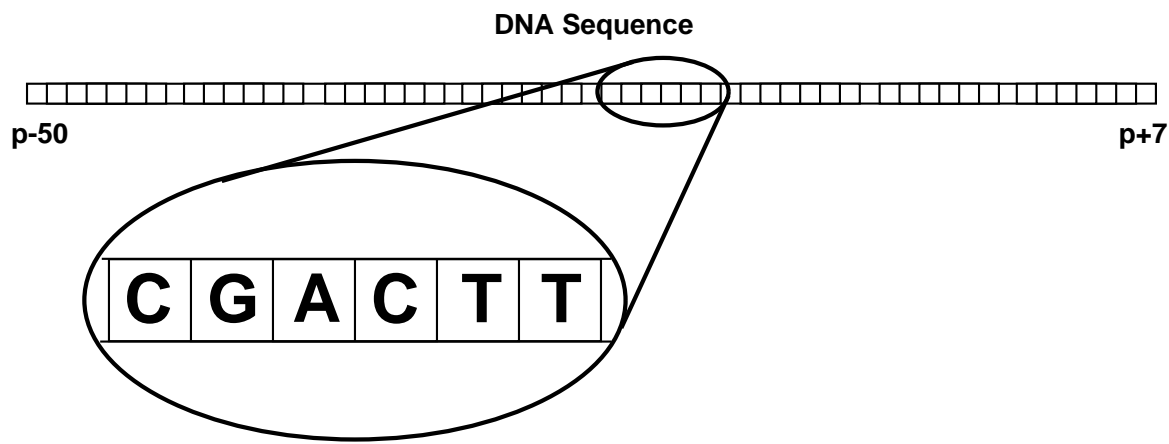

Figure 3: An instance in the promoter domain consists of a sequence of 57 nucleotides labeled from p-50 to p+7. Each nucleotide can take on the values A,G,C, or T representing adenine, guanine, cytosine, and thymine.

### 3.2 MIRO in the Promoter Domain

A MIRO-like system in the promoter domain would use the rules in Figure 4 to construct new high-level features for each DNA segment. Figure 6 shows an example of this. A DNA segment is shown from position p-38 through position p-30. The *minus_35* rules from the theory are also shown, and four new features (feat_minus35_A through feat_minus35_D) have been constructed for that DNA segment, one for each *minus_35* rule. The new features feat_minus35_A and feat_minus35_D both have the value 1 because the DNA fragment matches the first and fourth *minus_35* rules. Likewise, feat_minus35_B and feat_minus35_C both have the value 0 because the DNA fragment does not match the second and third rules. Furthermore, since the four *minus_35* rules are joined by disjunction, a new feature, feat_minus35_all, is created for the group that would have the value 1 because at least one of the *minus_35* rules matches.

New features would similarly be created for the *minus_10* rules and the *conformation* rules, and a standard induction algorithm could then be applied. We implemented a MIRO-like system; Figure 7 gives an example theory created by it. (Drastal's original MIRO used the candidate elimination algorithm (Mitchell, 1977) as its underlying induction algorithm. We used C4.5 (Quinlan, 1993).) As opposed to theory revision systems that incrementally modify the domain theory, MIRO has broken the theory down into its components and has fashioned these components into a new theory using a standard induction program. Thus MIRO has exhibited the *flexible structure* principle for this domain – it was not restricted in any way by the structure of the initial theory. Rather, MIRO exploited the strengths of standard induction to concisely characterize the training examples using the new features.

```
Promoters have a region where a protein (RNA polymerase) must make contact and
the helical DNA sequence must have a valid conformation so that the two pieces
of the contact region spatially align.  Prolog notation is used.

  promoter :- contact, conformation.

There are two regions "upstream" from the beginning of the gene at which the
RNA polymerase makes contact.

  contact  :- minus_35, minus_10.

The following rules describe the compositions of possible contact regions.

  minus_35 :- p-37=c, p-36=t, p-35=t, p-34=g, p-33=a, p-32=c.
  minus_35 :-         p-36=t, p-35=t, p-34=g,         p-32=c, p-31=a.
  minus_35 :-         p-36=t, p-35=t, p-34=g, p-33=a, p-32=c, p-31=a.
  minus_35 :-         p-36=t, p-35=t, p-34=g, p-33=a, p-32=c.

  minus_10 :- p-14=t, p-13=a, p-12=t, p-11=a, p-10=a, p-9=t.
  minus_10 :-         p-13=t, p-12=a,         p-10=a,         p-8=t.
  minus_10 :-         p-13=t, p-12=a, p-11=t, p-10=a, p-9=a, p-8=t.
  minus_10 :-                 p-12=t, p-11=a,                         p-7=t.

The following rules describe sequences that produce acceptable conformations.

  conformation :- p-47=c, p-46=a, p-45=a, p-43=t, p-42=t, p-40=a, p-39=c, p-22=g,
                  p-18=t, p-16=c, p-8=g,  p-7=c,  p-6=g,  p-5=c,  p-4=c,  p-2=c,
                  p-1=c.
  conformation :- p-45=a, p-44=a, p-41=a.
  conformation :- p-49=a, p-44=t, p-27=t, p-22=a, p-18=t, p-16=t, p-15=g, p-1=a.
  conformation :- p-45=a, p-41=a, p-28=t, p-27=t, p-23=t, p-21=a, p-20=a, p-17=t,
                  p-15=t, p-4=t.
```

Figure 4: The initial domain theory for recognizing promoters (from Towell and Shavlik).

A weakness Miro displays in this example is that it allows no flexibility of representation of the theory. The representation of the features constructed by Miro is basically the same all-or-none representation of the initial theory; either a DNA segment matched a rule, or it did not.

### 3.3 Either in the Promoter Domain

An Either-like system refines the initial promoter theory by dropping and adding conditions and rules. We simulated Either by turning off the M-of-N option in Neither and ran it in the promoter domain. Figure 8 shows the refined theory produced using a randomly selected training set of size 80. Because the initial promoter domain theory does not lend itself to revision through small, local changes, Either has only limited success.

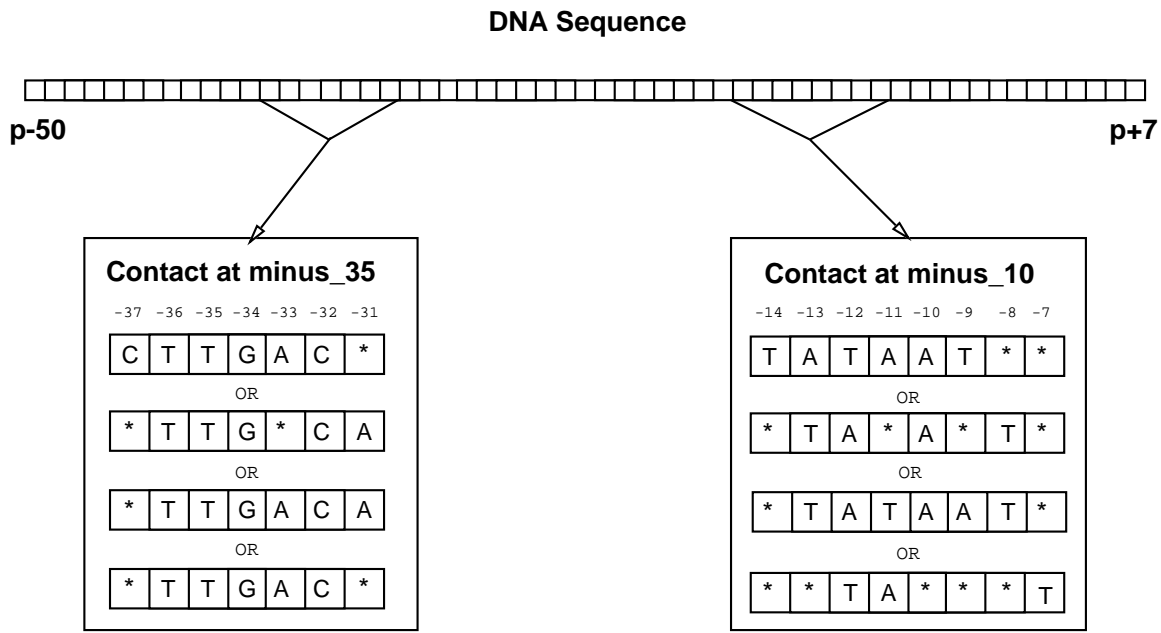

Figure 5: The contact portion of the theory. There are four possibilities for both the *minus_35* and *minus_10* portions of the theory. A "*" matches any nucleotide. The *conformation* portion of the theory is too spread out to display pictorially.

In this run, the program exhibited the second behavioral extreme discussed in Section 1; it entirely removed groups of rules and then tried to build new rules to replace what was lost. The *minus_10* and *conformation* rules have essentially been removed, and new rules have been added to the *minus_35* group. These new *minus_35* rules contain the condition p-12=t previously found in the *minus_10* group and the condition p-44=a previously found in the *conformation* group.

Either's behavior in this example is a direct result of its lack of flexibility of representation and flexibility of structure. It is difficult to transform the *minus_10* and *conformation* rules into something useful in their initial representation using Either's locally-acting operators. Either handles this by dropping these sets of rules, losing their knowledge, and attempting to rediscover the lost knowledge empirically. The end result of this loss of knowledge is lower than optimal accuracy shown later in Section 5.

### 3.4 Kbann in the Promoter Domain

Figure 9, modeled after a figure by Towell and Shavlik (1994), shows the setup of a Kbann network for the promoter theory. Each slot along the bottom represents one nucleotide in the DNA sequence. Each node at the first level up from the bottom embodies a single domain rule, and higher levels encode groups of rules with the final concept at the top. The links shown in the figure are the ones that are initially high-weighted. The net is next filled out to be fully connected with low-weight links. Backpropagation is then applied to refine the network's weights.

*A DNA segment fragment:*

...p-38=g, p-37=c, p-36=t, p-35=t, p-34=g, p-33=a, p-32=c, p-31=t, p-30=t ...

*The minus_35 group of rules and corresponding constructed features:*

| | | |
|---|---|---|
| minus_35 :- p-37=c, p-36=t, p-35=t, p-34=g, p-33=a, p-32=c. | feat_minus35_A = 1 |
| minus_35 :-      p-36=t, p-35=t, p-34=g,         p-32=c, p-31=a. | feat_minus35_B = 0 |
| minus_35 :-      p-36=t, p-35=t, p-34=g, p-33=a, p-32=c, p-31=a. | feat_minus35_C = 0 |
| minus_35 :-      p-36=t, p-35=t, p-34=g, p-33=a, p-32=c. | feat_minus35_D = 1 |

feat_minus35_all = (feat_minus35_A ∨ feat_minus35_B ∨ feat_minus35_C ∨ feat_minus35_D) = 1

Figure 6: An example of feature construction in a MIRO-like system. The constructed features for the first and fourth rules in the *minus_35* group are true (value = 1) because the DNA segment matches these rules. The constructed feature for the entire group, feat_minus35_all, is true because the four *minus_35* rules are joined by disjunction.

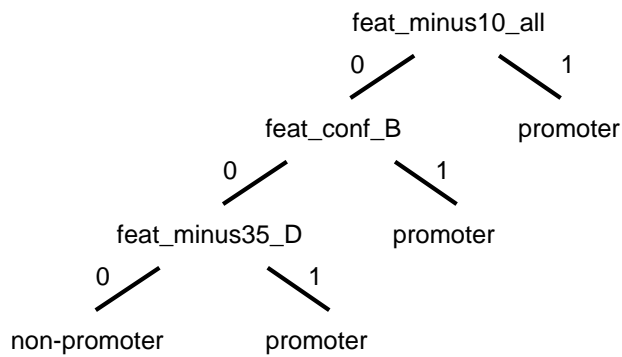

Figure 7: An example theory created by a MIRO-like system. A DNA segment is recognized as a promoter if it matches any of the *minus_10* rules, the second *conformation* rule, or the fourth *minus_35* rule.

The neural net representation is more appropriate for this domain than the propositional logic representation of the initial theory. It allows for a measurement of partial match by weighting the links in such a way that a subset of a rule's conditions are enough to surpass a node's threshold. It also allows for variable weightings of different parts of the theory; therefore, more predictive nucleotides can be weighted more heavily, and only slightly predictive nucleotides can be weighted less heavily. KBANN has only limited flexibility of structure. Because the refined network is the result of a series of incremental modifications in the initial network, a fundamental restructuring of the theory it embodies is unlikely. KBANN

```
promoter :- contact, conformation.

contact  :- minus_35, minus_10.

minus_35 :- p-35=t, p-34=g.
minus_35 :- p-36=t, p-33=a, p-32=c.
minus_35 :- p-36=t, p-32=c, p-50=c.
minus_35 :- p-34=g, p-12=t.
minus_35 :- p-34=g, p-44=a.
minus_35 :- p-35=t, p-47=g.

minus_10 :- true.

conformation :- true.
```

Figure 8: A revised theory produce by EITHER.

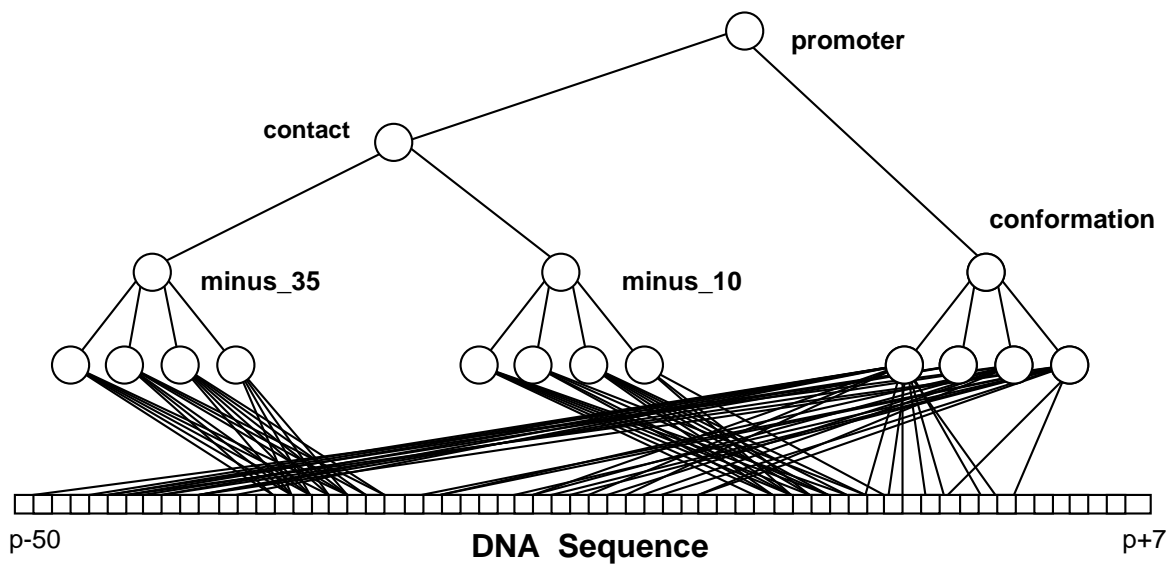

Figure 9: The setup of a KBANN network for the promoter theory.

is limited to finding the best network with the same fundamental structure imposed on it by the initial theory.

One of KBANN's advantages is that it uses a standard learning algorithm as its foundation. Backpropagation has been widely used and consequently improved by previous researchers. Theory refinement tools that are built from the ground up or use a standard tool only tangentially suffer from having to invent their own methods of handling standard problems such as overfit, noisy data, etc. A wealth of neural net experience and resources is available to the KBANN user; as neural net technology advances, KBANN technology will passively advance with it.

### 3.5 Neither-MofN in the Promoter Domain

Neither-MofN refines the initial promoter theory not only by dropping and adding conditions and rules but also by allowing conjunctive rules to be true if only a subset of their conditions are true. We ran Neither-MofN with a randomly selected training set of size 80, and Figure 10 shows a refined promoter theory produced. The theory expressed here with 9 M-of-N rules would require 30 rules using propositional logic, the initial theory's representation. More importantly, it is unclear how any system using the initial representation would reach the 30-rule theory from the initial theory. Thus the M-of-N representation adopted not only allows for the concise expression of the final theory but also facilitates the refinement process.

```
promoter :- 2 of ( contact, conformation ).

contact  :- 2 of ( minus_35, minus_10 ).

minus_35 :- 2 of ( p-36=t, p-35=t, p-34=g,          p-32=c, p-31=a ).
minus_35 :- 5 of ( p-36=t, p-35=t, p-34=g, p-33=a, p-32=c          ).

minus_10 :- 2 of (                  p-12=t, p-11=a,           p-7=t  ).
minus_10 :- 2 of (          p-13=t, p-12=a,        p-10=a,    p-8=t  ).
minus_10 :- 6 of ( p-14=t, p-13=a, p-12=t, p-11=a, p-10=a, p-9=t     ).
minus_10 :- 2 of (          p-13=t, p-12=a,        p-10=a,    p-34=g ).

conformation :- true.
```

Figure 10: A revised theory produced by Neither-MofN.

Neither-MofN displays flexibility of representation by allowing an M-of-N interpretation of the original propositional logic, but it does not allow for as fine-grained refinement as Kbann. Both allow for a measure of partial match, but Kbann could weight more predictive features more heavily. For example, in the $minus\_35$ rules, perhaps p-36=t is more predictive of a DNA segment being a promoter than p-34=g and therefore should be weighted more heavily. Neither-MofN simply counts the number of true conditions in a rule; therefore, every condition is weighted equally. Kbann's fine-grained weighting may be needed in some domains and not in others. It may actually be detrimental in some domains. An advanced theory revision system should offer a range of representations.

Like Kbann, Neither-MofN has only limited flexibility of structure. The refined theory is reached through a series of small, incremental modifications in the initial theory precluding a fundamental restructuring. Neither-MofN is therefore limited to finding the best theory with the same fundamental structure as the initial theory.

### 4. Theory-Guided Constructive Induction

In the first half of this section we present guidelines for *theory-guided constructive induction* that summarize the work discussed in Sections 2 and 3. The remainder of the section presents an algorithm that instantiates these guidelines. We evaluate the algorithm in Section 5.

### 4.1 Guidelines

The following guidelines are a synthesis of the strengths of the previously discussed related work.

- As in MIRO, new features should be constructed using components of the domain theory. These new features are combinations of existing features, and a final theory is created by applying a standard induction algorithm to the training examples described using the new features. This allows knowledge to be gleaned from the initial theory without forcing the final theory to conform to the initial theory's backbone structure. It takes full advantage of the domain theory by building high-level features from the original low-level features. It also takes advantage of a strength of standard induction — building concise theories having high predictive accuracy when the target concept *can* be concisely expressed using the given features.

- As in EITHER, the constructed features should be modifiable by various operators that act locally, such as adding or dropping conjuncts from a constructed feature.

- As in KBANN and NEITHER-MOFN, the representation of the constructed features need not be the exact representation in which the initial theory is given. For example, the initial theory may be given as a set of rules written in propositional logic. A new feature can be constructed for each rule, but it need not be a boolean feature telling whether all the conditions are met; for example it may be a count of how many conditions of that rule are met. This allows the final theory to be formed and expressed in a representation that is more suitable than the representation of the initial theory.

- Like GRENDEL, a complete system should offer a library of interpreters allowing the domain theory to be translated into a range of representations with differing adaptability. One interpreter might emulate MIRO strictly translating a domain theory into boolean constructed features. Another interpreter might construct features that count the number of satisfied conditions of the corresponding component of the domain theory thus providing a measure of partial match. Still another interpreter might construct features that are weighted sums of the satisfied conditions. The weights could be refined empirically by examining a set of training examples. Thus the most appropriate amount of expressive power can be applied to a given problem without incurring unnecessary expense.

### 4.2 A Specific Interpreter

This section describes an algorithm which is a limited instantiation of the guidelines just described. The algorithm is intended as a demonstration of the distillation and synthesis of the principles embodied in previous landmark systems. It contains a main module, TGCI described in Figure 12, and a specific interpreter, TGCI1 described in Figure 11. The main module TGCI redescribes the training and testing examples by calling TGCI1 and then applies C4.5 to the redescribed examples (just as MIRO applied the candidate elimination algorithm to examples after redescribing them). TGCI1 can be viewed as a single interpreter from a potential GRENDEL-like toolbox. It takes as input a single example and a domain theory expressed as an AND/OR tree such as the one shown in Figure 13. It returns a new vector of features for that example that measure the partial match of the example to the theory. Thus it creates new features from components of the domain theory as in MIRO, but because it measures partial match, it allows flexibility in representing the information contained in the initial theory as in KBANN and NEITHER-MOFN. One aspect of the guidelines in 4.1 that does not appear in this algorithm is EITHER's locally acting operators such as adding and dropping conditions from a portion of the theory. The following two paragraphs explain in more detail the workings of TGCI1 and TGCI respectively.

> *Given:* An example $E$ and a domain theory with root node $R$. The domain theory is an AND/OR/NOT tree in which the leaves are conditions which can be tested to be *true* or *false*.
>
> *Return:* A pair $P = (F, \mathcal{F})$ where $F$ is the top feature measuring the partial match of $E$ to the whole domain theory, and $\mathcal{F}$ is a vector of new features measuring the partial match of $E$ to various parts and subparts of the domain theory.

1. If $R$ is a directly testable condition, return P=(1,<>) if $R$ is *true* for $E$ and P=(-1,<>) if $R$ is *false* for $E$.

2. $n$ = the number of children of $R$

3. For each child $R_j$ of $R$, call TGCI1($R_j$,$E$) and store the respective results in $P_j = (F_j, \mathcal{F}_j)$.

4. If the major operator of $R$ is OR, $F_{new} = MAX(F_j)$.
   Return $P = (F_{new}, concatenate(<F_{new}>, \mathcal{F}_1, \mathcal{F}_2, ..., \mathcal{F}_n))$.

5. If the major operator of $R$ is AND, $F_{new} = (\sum_{j=1}^{n} F_j)/n$.
   Return $P = (F_{new}, concatenate(<F_{new}>, \mathcal{F}_1, \mathcal{F}_2, ..., \mathcal{F}_n))$.

6. If the major operator of $R$ is NOT, $F_{new} = -1 * F_1$.
   Return $P = (F_{new}, \mathcal{F}_1)$.

Figure 11: The TGCI1 algorithm

The TGCI1 algorithm, given in Figure 11, is recursive. Its inputs are an example $E$ and a domain theory with root node $R$. It ultimately returns a redescription of $E$ in the form of a vector of new features $\mathcal{F}$. It also returns a value $F$ called the *top feature* which is used in intermediate calculations described below. The base case occurs if the domain theory is a single leaf node (i.e., $R$ is a simple condition). In this case (Line 1), TGCI1 returns the top feature 1 if the condition is *true* and -1 if the condition is *false*. No new features are returned in the base case because they would simply duplicate the existing features. If the domain theory is not a single leaf node, TGCI1 recursively calls itself on each of $R$'s children (Line 3). When a child of $R$, $R_j$, is processed, it returns a vector of new features $\mathcal{F}_j$ (which measures the partial match of the example to the $j$th child of $R$ and its various subparts). It also returns the top feature $F_j$ which is included in $\mathcal{F}_j$ but is marked as special because it measures the partial match of the example to the *whole* of the $j$th child of $R$. If there are $n$ children, the result of Line 3 is $n$ vectors of new features, $\mathcal{F}_1$ to $\mathcal{F}_n$, and $n$ top features, $F_1$ to $F_n$. If the operator at node $R$ is OR (Line 4), then $F_{new}$, the new feature created for that node, is the maximum of $F_j$. Thus $F_{new}$ measures how closely the *best* of $R$'s children come to having its conditions met by the example. The vector of new features returned in this case is a concatenation of $F_{new}$ and all the new features from $R$'s children. If the operator at node $R$ is AND (Line 5), then $F_{new}$ is the average of $F_j$. Thus $F_{new}$ measures how closely all of $R$'s children as a group come to having their conditions met by the example. The vector of new features returned in this case is again a concatenation of $F_{new}$ and all the new features from $R$'s children. If the operator at node $R$ is NOT (Line 6), $R$ should only have one child, and $F_{new}$ is $F_1$ negated. Thus $F_{new}$ measures the extent to which the conditions of $R$'s child are *not* met by the example.

> *Given:* A set of training examples $E_{train}$, a set of testing examples $E_{test}$, and a domain theory with root node $R$.
>
> *Return:* Learned concept and accuracy on testing examples.
>
> 1. For each example $E_i \in E_{train}$, call TGCI1($R,E_i$) which returns $P_i = (F_i, \mathcal{F}_i)$. $E_{train-new} = \{\mathcal{F}_i\}$.
>
> 2. For each example $E_i \in E_{test}$, call TGCI1($R,E_i$). which returns $P_i = (F_i, \mathcal{F}_i)$. $E_{test-new} = \{\mathcal{F}_i\}$.
>
> 3. Call C4.5 with training examples $E_{train-new}$ and testing examples $E_{test-new}$. Return decision tree and accuracy on $E_{test-new}$.

Figure 12: The TGCI algorithm

If TGCI1 is called twice with two different examples but with the same domain theory, the two vectors of new features will be the same size. Furthermore, corresponding features measure the match of corresponding parts of the domain theory. The TGCI main module in Figure 12 takes advantage of this by creating redescribed example sets from the input example sets. Line 1 redescribes each example in the training set producing a new training set. Line 2 does the same for the testing set. Line 3 runs the standard induction program C4.5 on these redescribed example sets. The returned decision tree can be easily interpreted by examining which new features were used and what part of the domain theory they correspond to.

### 4.3 TGCI1 Examples

As an example of how the TGCI1 interpreter works, consider the toy theory shown in Figure 13. TGCI1 redescribes the input example by constructing a new feature for each node in the input theory. Consider the situation where the input example matches conditions A, B, and D but not C and E. When TGCI1 evaluates the children of Node 6, it gets the values $F_1 = 1$, $F_2 = 1$, $F_3 = -1$, $F_4 = 1$, and $F_5 = -1$. Since the operator at Node 6 is AND, $F_{new}$ is the average of the values received from the children, 0.20 $((1 + 1 + (-1) + 1 + (-1))/5 = 0.20)$. Likewise, if condition G matchs but not F and H, $F_{new}$ for Node 5 will have the value 0.33 $(-1 * ((1 + (-1) + (-1))/3))$ because two of three matching conditions at Node 7 give the value $-0.33$, and this is negated by the NOT at Node 5. Since Node 2 is a disjunction, its new feature measures the best partial match of its two children and has the value 0.33 (MAX(0.20,0.33)), and so on.

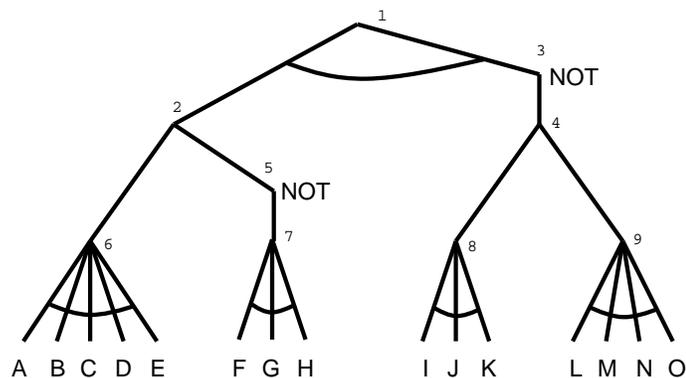

Figure 13: An example theory in the form of an AND/OR tree that might be used by the interpreter to generate constructed features.

Figure 14 shows how TGCI1 redescribes a particular DNA segment using the $minus\_35$ rules of the promoter theory. A partial DNA segment is shown along with the four $minus\_35$ rules and the new feature constructed for each rule (We have given the new features names here to simplify our illustration). For the first rule, four of the six nucleotides match; therefore, for that DNA segment feat_minus35_A has the value 0.33 $((1+1+1+1+(-1)+(-1))/6)$. For the second rule, four of the five nucleotides match; therefore, feat_minus35_B has the value 0.60. Because these and the other two $minus\_35$ rules are joined by disjunction in the original domain theory, feat_minus35_all, the new feature constructed for this group, takes the maximum value of its four children; therefore, feat_minus35_all has the value 0.60 because feat_minus35_B has the value 0.60, the highest in the group. Intuitively, feat_minus35_all represents the best partial match of this grouping — the extent to which the disjunction is partially satisfied. The results of running TGCI1 on each DNA sequence is a set of redescribed training examples. Each redescribed example has a value for feat_minus35_A through feat_minus35_D, feat_minus35_all, and all other nodes in the promoter domain theory. The training set is essentially redescribed using a new feature vector derived from information contained in the domain theory. In this form, any off-the-shelf induction program can be applied to the new example set.

Anomalous situations can be created in which TGCI1 gives a "good score" to a seemingly bad example and a bad score to a good example. Situations can also be created where logically equivalent theories give different scores for a single example. These occur because

*A DNA segment fragment:*

...p-38=g, p-37=c, p-36=t, p-35=t, p-34=g, p-33=c, p-32=a, p-31=a, p-30=t ...

*The minus_35 group of rules and corresponding constructed features:*

| | | |
|---|---|---|
| minus_35 :- p-37=c, p-36=t, p-35=t, p-34=g, p-33=a, p-32=c. | | feat_minus35_A = 0.33 |
| minus_35 :-          p-36=t, p-35=t, p-34=g,          p-32=c, p-31=a. | | feat_minus35_B = 0.60 |
| minus_35 :-          p-36=t, p-35=t, p-34=g, p-33=a, p-32=c, p-31=a. | | feat_minus35_C = 0.33 |
| minus_35 :-          p-36=t, p-35=t, p-34=g, p-33=a, p-32=c. | | feat_minus35_D = 0.20 |

feat_minus35_all = max(feat_minus35_A, feat_minus35_B, feat_minus35_C, feat_minus35_D) = 0.60

Figure 14: An example of how TGCI1 generates constructed features from a portion of the promoter domain theory and a DNA segment. Four of the conditions in the first *minus_35* rule match the DNA segment; therefore, the constructed feature for that rule has the value 0.33 $((1 + 1 + 1 + 1 + (-1) + (-1))/6)$. Feat_minus35_all, the new feature for the entire *minus_35* group takes the maximum value of its children thus embodying the best partial match of the group.

TGCI1 is biased to favor situations where more matched conditions of an AND is desirable, but more matched conditions of an OR is not necessarily better. Eliminating these anomalies would remove this bias.

## 5. Experiments and Analysis

This section presents the results of applying theory-guided constructive induction to three domains: the promoter domain (Harley et al., 1990), the primate splice-junction domain (Noordewier, Shavlik, & Towell, 1992), and the gene identification domain (Craven & Shavlik, 1995). In each case the TGCI1 interpreter was applied to the domain's theory and examples in order to redescribe the examples using new features. Then C4.5 (Quinlan, 1993) was applied to the redescribed examples.

### 5.1 The Promoter Domain

Figure 15 shows a learning curve for theory-guided constructive induction in the promoter domain accompanied by curves for EITHER, LABYRINTH$_K$, KBANN, and NEITHER-MOFN. Following the methodology described by Towell and Shavlik [1994], the set of 106 examples was randomly divided into a training set of size 80 and a testing set of size 26. A learning curve was created by training on subsets of the training set of size 8, 16, 24, ...72, 80, using the 26 examples for testing. The curves for EITHER, LABYRINTH$_K$, and KBANN were taken from Ourston and Mooney (1990), Thompson, Langley, and Iba (1991), and Towell and Shavlik (1994) respectively and were obtained by a similar methodology[1]. The curve for TGCI is the average of 50 independent random data partitions and is given along with 95% confidence ranges. The NEITHER-MOFN program was obtained from Ray Mooney's group and was used in generating the NEITHER-MOFN curve using the same 50 data partitions as were used for TGCI[2].

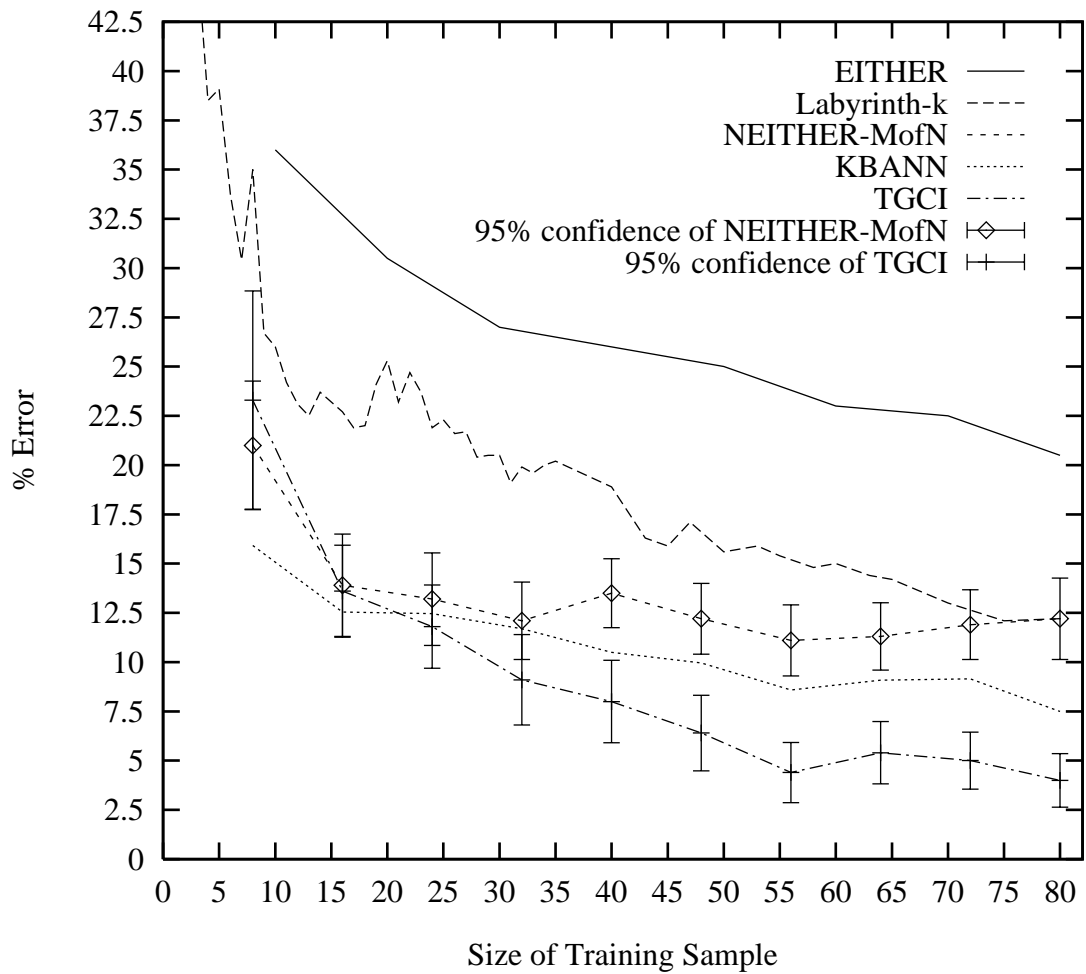

Figure 15: Learning curves for theory-guided constructive induction and other systems in the promoter domain.

TGCI showed improvement over EITHER and LABYRINTH$_K$ for all portions of the curve and also performed better than KBANN and NEITHER-MOFN for all except the smallest training sets. Confidence intervals were not available for EITHER, LABYRINTH$_K$, and

---

1. EITHER used a testing set of size 25 and did not use the *conformation* portion of the domain theory. The testing set in LABYRINTH$_K$ always consisted of 13 promoters and 13 non-promoters.
2. Baffes and Mooney (1993) report a slightly better learning curve for NEITHER-MOFN than we obtained, but after communication with Paul Baffes, we think the difference is caused by the random selection of data partitions.

KBANN, but in a pairwise comparison with NEITHER-MOFN, the improvement of TGCI was significant at the 0.0005 level of confidence for training sets of size 48 and larger.

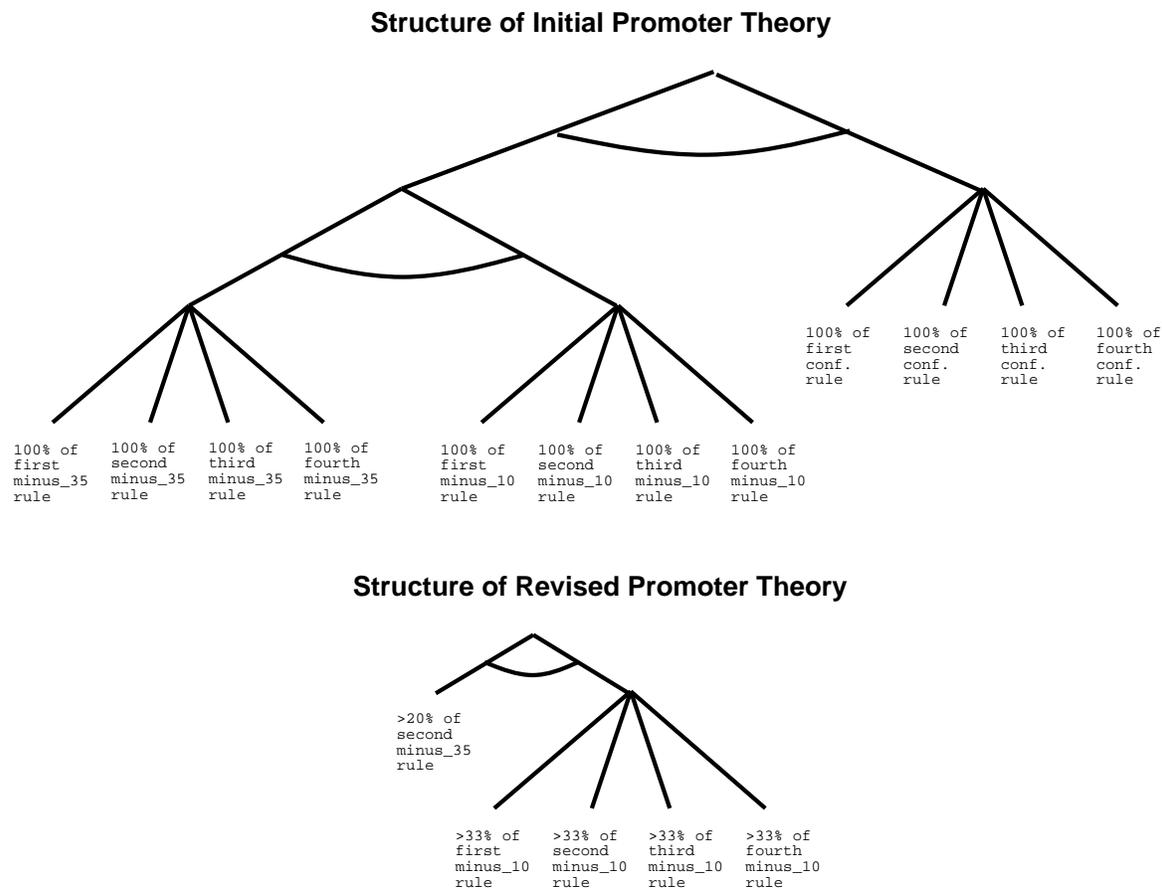

Figure 16: The revised theory produced by theory-guided constructive induction has borrowed substructures from the initial theory, but as a whole has not been restricted by its structure.

Figure 16 compares the initial promoter theory with a theory created by TGCI. Reasons for TGCI's improvement can be inferred from this figure. TGCI has extracted the components of the original theory that are most helpful and restructured them into a more concise theory. Neither KBANN nor NEITHER-MOFN facilitates this radical extraction and restructuring. As seen in the leaf nodes, the new theory also measures the partial match of an example to components of the original theory. This aspect is similar in KBANN and NEITHER-MOFN.

Part of TGCI's improvement over KBANN and NEITHER-MOFN may be due to a *knowledge/bias conflict* in the latter two systems, a situation where revision biases conflict with knowledge in such a way as to undo some of the knowledge's benefits. This can occur whenever detailed knowledge is opened up to revision using a set of examples. The revision is not guided *only* by the examples but rather by the examples as interpreted by a set of algorithmic biases. Biases that are useful in the absence of knowledge may undo good knowledge when improperly applied. Yet these biases developed and perfected for pure induction are often unquestioningly applied to the revision of theories. The biases governing the dropping of conditions in Neither-MofN and reweighting conditions in Kbann may be neutralizing the promoter theory's potential. We speculate this because we conducted some experiments that allowed bias-guided dropping and adding of conditions within Tgci. We found that these techniques actually reduced accuracy in this domain.

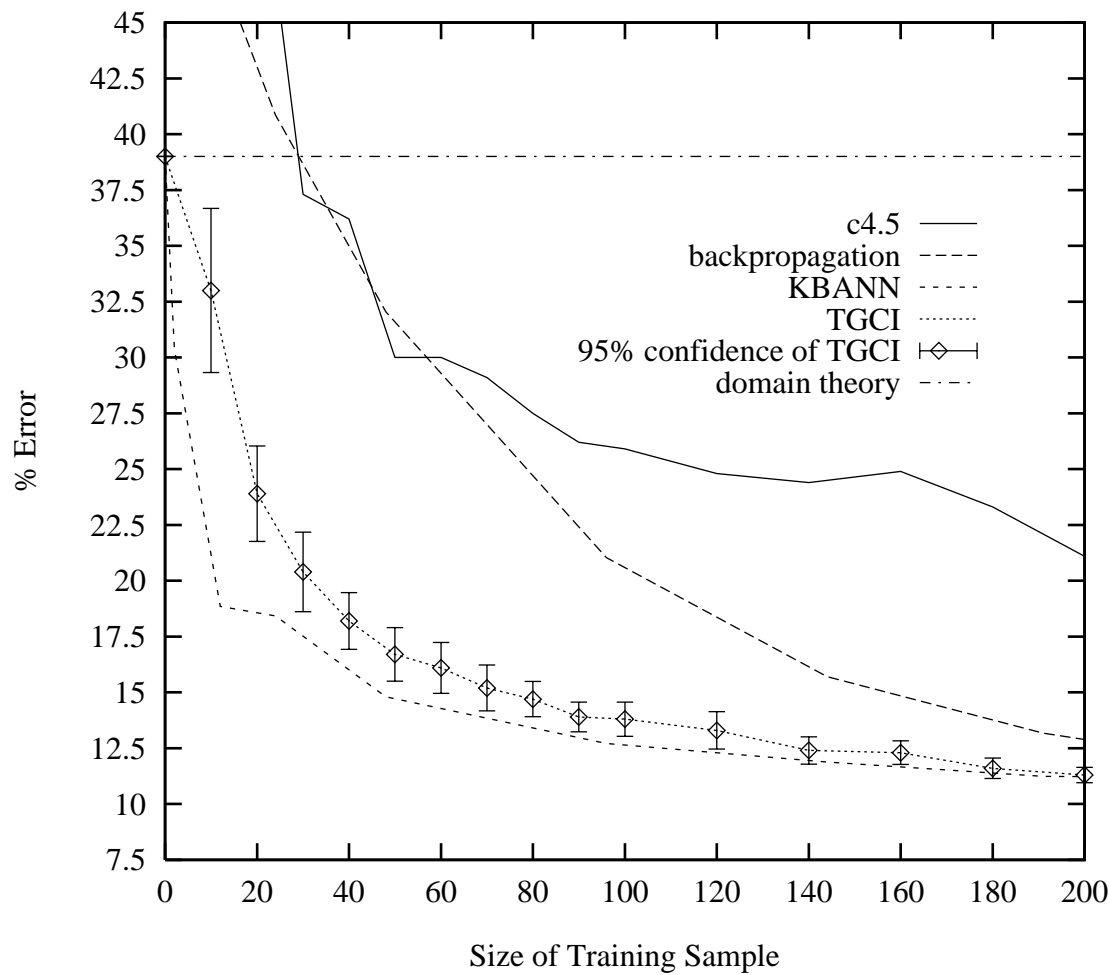

Figure 17: Learning curves for Tgci and other systems in the primate splice-junction domain.

## 5.2 The Primate Splice-junction Domain

The primate splice-junction domain (Noordewier et al., 1992) involves analyzing a DNA sequence and identifying boundaries between introns and exons. Exons are the parts of a DNA sequence kept after splicing; introns are spliced out. The task then involves placing a given boundary into one of three classes: an intron/exon boundary, an exon/intron boundary, or neither. An imperfect domain theory is available which has a 39.0% error rate on the entire set of available examples.

Figure 17 shows learning curves for C4.5, backpropagation, KBANN, and TGCI in the primate splice-junction domain. The results for KBANN and backpropagation were taken from Towell and Shavlik (1994). The curves for plain C4.5 and the TGCI algorithm were created by training on sets of size 10,20,30,...90,100,120,...200 and testing on a set of size 800. The curves for C4.5 and TGCI are the average of 40 independent data partitions. No comparison was made with NEITHER-MOFN because the implementation we obtained could handle only two-class concepts. For training sets larger than 200, KBANN, TGCI, and backpropagation all performed similarly.

The accuracy of TGCI appears slightly worse than that of KBANN but perhaps not significantly. KBANN's advantage over TGCI is its ability to assign fine-grained weightings to individual parts of a domain theory. TGCI's advantage over KBANN is its ability to more easily restructure the information contained in a domain theory. We speculate that KBANN's capability to assign fine-grained weights outweighted its somewhat rigid structuring of this domain theory. Theory-guided constructive induction has an advantage of speed over KBANN because C4.5, its underlying learner, runs much more quickly than backpropagation, KBANN's underlying learning algorithm.

### 5.3 The Gene Identification Domain

The gene identification domain (Craven & Shavlik, 1995) involves classifying a given DNA segment as a coding sequence (one that codes a protein) or a non-coding sequence. No domain theory was available in the gene identification domain; therefore, we created an artificial domain theory using the information that organisms may favor certain nucleotide triplets over others in gene coding. The domain theory embodies the knowledge that a DNA segment is likely to be a gene coding segment if its triplets are coding-favoring triplets or if its triplets are not noncoding-favoring triplets. The decision of which triplets were coding-favoring, which were noncoding-favoring, and which favored neither, was made empirically by analyzing the makeup of 2500 coding and 2500 noncoding sequences. The specific artificial domain theory used is described in Online Appendix 1.

Figure 18 shows learning curves for C4.5 and TGCI in the gene identification domain. The original domain theory yields 31.5% error. The curves were created by training on example sets of size 50,200,400,...2000 and testing on a separate example set of size 1000. The curves are the average of 40 independent data partitions.

Only a partial curve is given for NEITHER-MOFN because it became prohibitively slow for larger training sets. In the promoter domain where training sets were smaller than 100, TGCI and NEITHER-MOFN ran at comparable speeds (approximately 10 seconds on a Sun4 workstation). In this domain TGCI ran in approximately 2 minutes for larger training sets. NEITHER-MOFN took 21 times as long as TGCI on training sets of size 400, 69 times as long for size 800, and 144 times as long for size 1200. Consequently, NEITHER-MOFN's curve only extends to 1200 and only represents five randomly selected data partitions. For these reasons, a solid comparison of NEITHER-MOFN and TGCI cannot be made from these curves, but it appears that TGCI's accuracy is slightly better. We speculate that NEITHER-

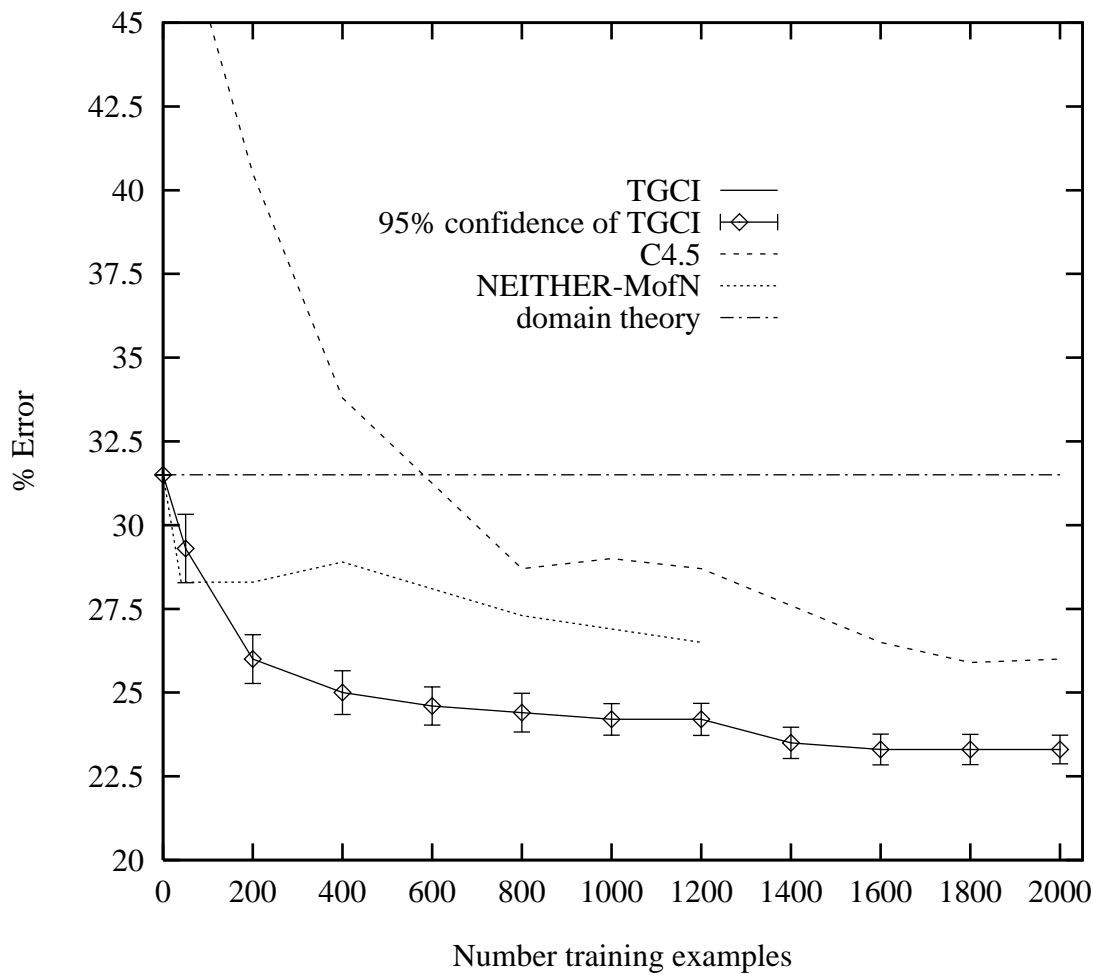

Figure 18: Learning curves for TGCI and other systems in the gene identification domain.

MOFN's slightly lower accuracy is partially due to the fact that it revises the theory to correctly classify *all* the training examples. The result is a theory which likely overfits the training examples. TGCI does not need to explicitly avoid overfit because this is handled by its underlying learner.

### 5.4 Summary of Experiments

Our goal in this paper has *not* been to present a new technique but rather to understand the behavior of landmark systems, distill their strengths, and synthesize them into a simple system, TGCI. The evaluation of this algorithm shows that its accuracy roughly matches or exceeds that of its predecessors. In the promoter domain, TGCI showed sizable improvement over many published results. In the splice-junction domain, TGCI narrowly falls short of KBANN's accuracy. In the gene identification domain, TGCI outperforms NEITHER-MOFN. In all these domains TGCI greatly improves on the original theory alone and C4.5 alone.

TGCI is faster than its closest competitors. TGCI runs as much as 100 times faster than NEITHER-MOFN on large datasets. A strict quantitative comparison of the speeds of TGCI and KBANN was not made because 1) backpropagation is known to be much slower than decision trees (Mooney, Shavlik, Towell, & Gove, 1989), 2) KBANN uses multiple hidden layers which makes its training time even longer (Towell & Shavlik, 1994), and 3) Towell and Shavlik (1994) point out that each run of KBANN must be made multiple times with different initial random weights, whereas a single run of TGCI is sufficient.

Overall, our experiments support two claims of this paper: First, the accuracy of TGCI substantiates our delineation of system strengths in terms of flexible theory representation and flexible theory structure, since this characterization is the basis for this algorithm's design. Second, TGCI's combination of speed and accuracy suggest that unnecessary computational complexity can be avoided in synthesizing the strengths of landmark systems. In the following section we take a closer look at the strengths of theory-guided constructive induction.

## 6. Discussion of Strengths

Below a number of strengths of theory-guided constructive induction are discussed within the context of the TGCI algorithm used in our experiments.

### 6.1 Flexible Representation

As discussed in Section 1, for many domains the representation most appropriate for an initial theory may not be most appropriate for a refined theory. Because theory-guided constructive induction allows the translation of the initial theory into a different representation, it can accommodate such domains. In the experiments in this paper a representation was needed which allowed for a measurement of partial match to the domain theory. TGCI1 accomplished this by simply counting the matching features and propagating this information up the theory appropriately. EITHER and LABYRINTH$_K$ do not easily afford this measure of partial match and therefore are more appropriate for problems in which the best representation of the final theory is the same as that of the initial theory. KBANN allows a finer-grained measurement of partial match than both NEITHER-MOFN and our work, but a price is paid in computational complexity. The theory-guided constructive induction framework allows for a variety of potential tools with varying degrees of granularity of partial match, although just one tool is used in our experiments.

### 6.2 Flexible Structure

As discussed in Section 2.5, a strength of existing induction programs is fashioning a concise and highly predictive description of a concept when the target concept *can* be concisely described with the given features. Consequently, the value of a domain theory lies *not* in its overall structure. If the feature language is sufficient, any induction program can build a good overall theory structure. Instead, the value of a domain theory lies in the information it contains about how to redescribe examples using high-level features. These high-level features facilitate a concise description of the target concept. Systems such as EITHER and NEITHER-MOFN that reach a final theory through a series of modifications in the initial theory hope to gain something by keeping the theory's overall structure intact. If the initial theory is sufficiently close to an accurate theory, this method works, but often clinging to the structure hinders full exploitation of the domain theory. Theory-guided constructive induction provides a means of fully exploiting both the information in the domain theory *and* the strengths of existing induction programs. Figure 16 in Section 5.1 gives a comparison of the structure of the initial promoter theory to the structure of a revised theory produced by theory-guided constructive induction. Substructures have been borrowed, but the revised theory as a whole has been restructured.

### 6.3 Use of Standard Induction as an Underlying Learner

Because theory-guided constructive induction uses a standard induction program as its underlying learner, it does not need to reinvent solutions to overfit avoidance, multi-class concepts, noisy data, etc. Overfit avoidance has been widely studied for standard induction, and many standard techniques exist. Any system which modifies a theory to accommodate a set of training examples must also address the issue of overfit to the training examples. In many theory revision systems existing overfit avoidance techniques cannot be easily adapted, and the problem must be addressed from scratch. Theory-guided constructive induction can take advantage of the full range of previous work in overfit avoidance for standard induction.

When multiple theory parts are available for multi-class concepts, the interpreter is run on the multiple theory parts, and the resulting new feature sets are combined. The primate splice-junction domain presented in Section 5.2 has three classes: intron/exon boundaries, exon/intron boundaries, and neither. Theories are given for both intron/exon and exon/intron. Both theories are used to create new features, and then all new features are concatenated together for learning. Interpreters such as TGCI1 also trivially handle negation in a domain theory.

### 6.4 Use of Theory Fragments

Theory-guided constructive induction is not limited to using full domain theories. If only part of a theory is available, this can be used. To demonstrate this, three experiments were run in which only fragments of the promoter domain theory were used. In the first experiment, only the four *minus_35* rules were used. Five features were constructed — one feature for each rule and then an additional feature for the group. Similar experiments were run for the *minus_10* group and the *conformation* group.

Figure 19 gives learning curves for these three experiments along with curves for the entire theory and for no theory (C4.5 using the original features). Although the *conformation* portion of the theory gives no significant improvement over C4.5, both the *minus_35* and *minus_10* portions of the theory give significant improvements in performance. Thus even partial theories and theory fragments can be used by theory-guided constructive induction to yield sizable performance improvements.

The use of theory fragments should be explored as a means of evaluating the contribution of different parts of a theory. In Figure 19, the *conformation* portion of the theory is shown to yield no improvement. This could signal a knowledge engineer that the knowledge that should be conveyed through that portion of the theory is not useful to the learner in its present form.

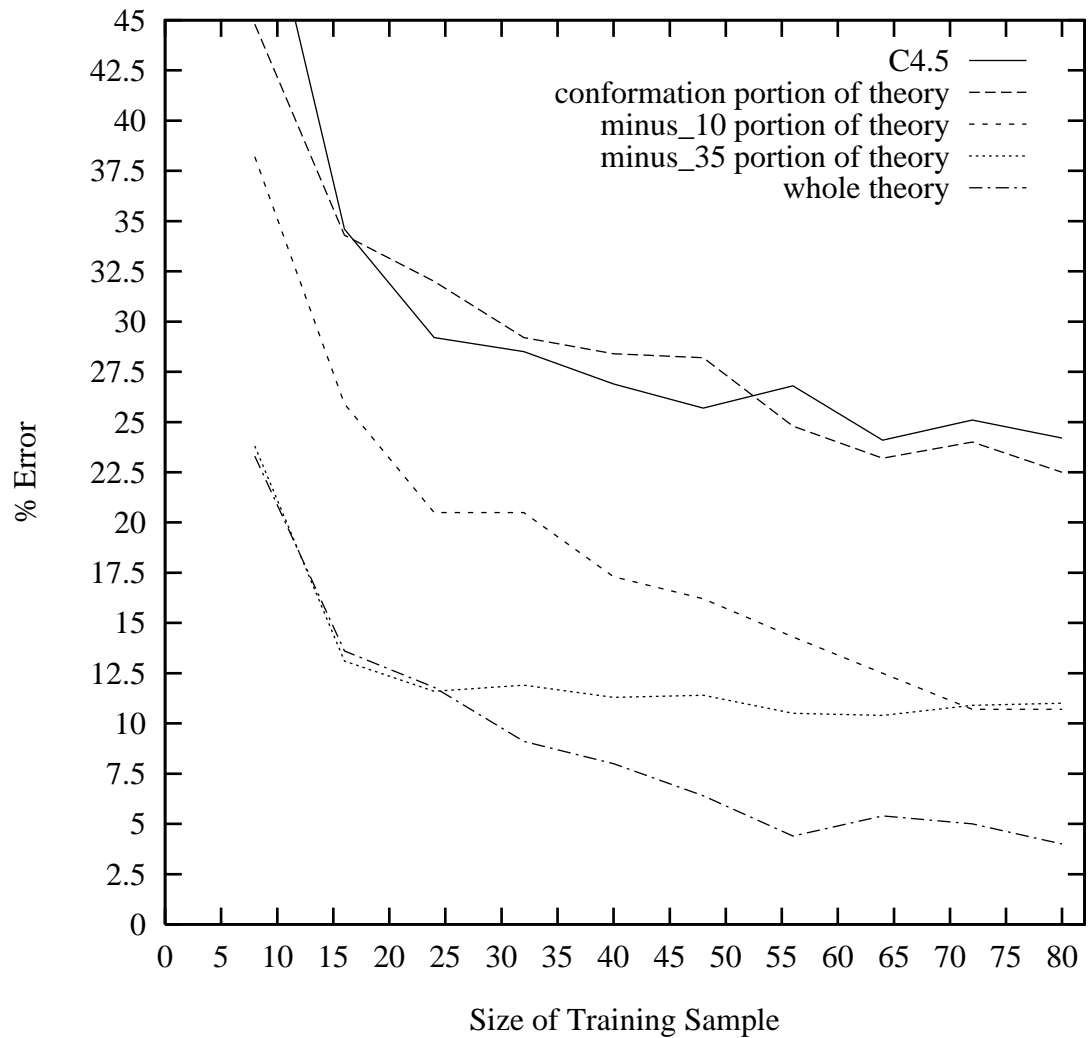

Figure 19: Learning curves for theory-guided constructive induction with only fragments of the promoter domain theory. The *minus_35* portion of the theory, the *minus_10* portion of the theory, and the *conformation* portion of the theory were used separately in feature construction. Curves are also given for the full theory and for C4.5 alone for comparison.

### 6.5 Use of Multiple Theories

Theory-guided constructive induction can use multiple competing and even incompatible domain theories. If multiple theories exist, theory-guided constructive induction provides a natural means of integrating them in such a way as to extract the best from all theories. TGCI1 would be called for each input theory producing new features. Next, all the new features are simply pooled together and the induction program selects from among them in fashioning the final theory. This is seen on a very small scale in the promoter domain.

In Figure 4 some *minus_35* rules subsume other *minus_35* rules. According to the entry in the UCI Database, this is because "the biological evidence is inconclusive with respect to the correct specificity." This is handled by simply using all four possibilities, and selection of the most useful knowledge is left to the induction program.

TGCI could also be used to evaluate the contributions of competing theories just as it was used to evaluate theory fragments above. A knowledge engineer could use this evaluation to guide his own revision and synthesis of competing theories.

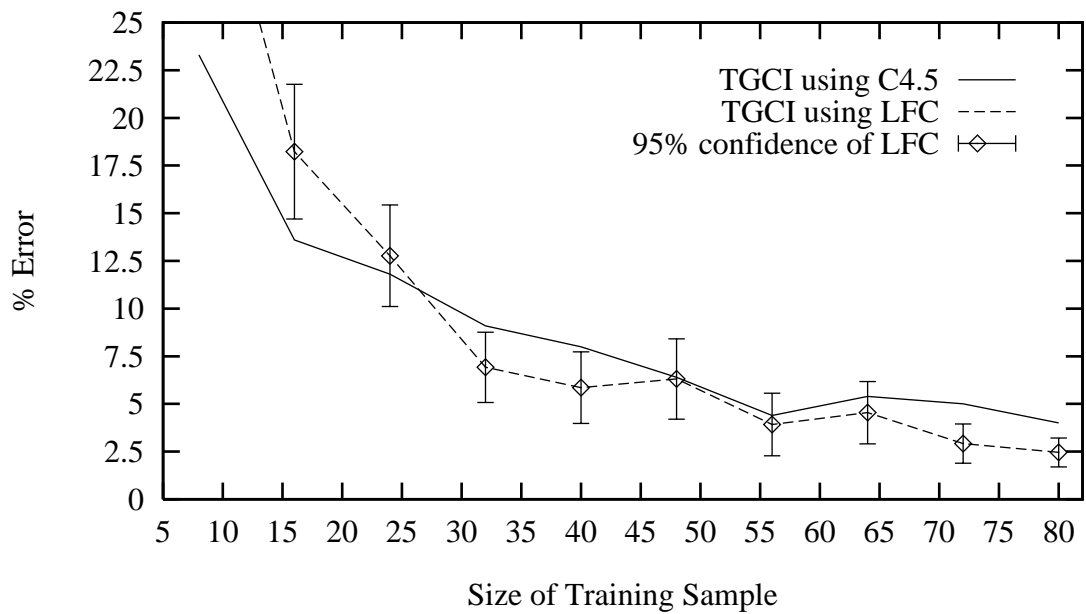

Figure 20: Theory-guided constructive induction with LFC and C4.5 as the underlying learning system. Theory-guided constructive induction can use any inductive learner as its underlying learning component. Therefore, more sophisticated underlying induction programs can further improve accuracy.

### 6.6 Easy Adoption of New Techniques

Since theory-guided constructive induction can use any standard induction method as its underlying learner, as improvements are made in standard induction, theory-guided constructive induction passively improves. To demonstrate this, tests were also run with LFC (Ragavan & Rendell, 1993) as the underlying induction program. LFC is a decision tree learner that performs example-based constructive induction by looking ahead at combinations of features. Characteristically, LFC improves accuracy for a moderate number of examples. Figure 20 shows the resulting learning curve along with the C4.5 TGCI curve. Both curves are the average of 50 separate runs with the same data partitions used for each program. In a pairwise comparison the improvement of LFC over C4.5 was significant at the 0.025 level of confidence for training sets of size 72 and 80. More sophisticated underlying induction programs can further improve accuracy.

## 7. Testing the Limits of TGCI

The purpose of this section is to explore the performance of theory-guided constructive induction on theory revision problems ranging from easy to difficult. In easy problems the underlying concept embodied in the training and testing examples matches the domain theory fairly closely; therefore, the examples themselves match the domain theory fairly closely. In difficult problems the underlying concept embodied in the examples does not match the domain theory very well so the examples do not either. Although many other factors determine the difficulty of an individual problem, this aspect is an important component and worth exploring. Our experiment in this section is intended to relate ranges of difficulty to the amount of improvement produced by TGCI.

Since a number of factors affect problem difficulty we chose that the theory revision problems for the experiment should all be variations of a single problem. By doing this we are able to hold all other factors constant and vary the closeness of match to the domain theory. Because we wanted to avoid totally artificial domains, we chose to start with the promoter domain and create "new" domains by perverting the example set.

These "new" domains were created by perverting the examples in the original promoter problem to either *more* closely match the promoter domain theory or *less* closely match the promoter domain theory. Only the positive examples were altered. For example, one domain was created with 30% fewer matches to the domain theory than the original promoter domain as follows: Each feature value in a given example was examined to see if it matched part of the theory. If so, with a 30% probability, it was randomly reassigned a new value from the set of possible values for that feature. The end result is a set of examples with 30% fewer matches to the domain theory than the original example set[3]. For our experiment new domains such as this were created with 10%, 30%, 60%, and 90% fewer matches.

For some features, multiple values may match the theory because different disjuncts of the theory specify different values for a single feature. For example, referring back to Figure 4, feature $p$-12 matches two of the $minus\_10$ rules if it has the value $a$ and another two rules if it has the value $t$. So a single feature might accidentally match one part of a theory when in fact the example as a whole more closely matches another part of the theory. For cases such as these, true matches were separated from accidental matches by examining which part of the theory most clearly matched the example as a whole and expecting a match from that part of the theory.

New domains that *more* closely matched the theory were created in a similar manner. For example, a domain was created with 30% fewer *mis*matches to the domain theory than the original promoter domain as follows: Each feature value in a given example was examined to see if it matched its corresponding part of the theory. If not, with a 30% probability, it was reassigned a value that matched the theory. The end result is a set of examples in which 30% of the mismatches with the domain theory are eliminated. For our experiment new domains such as this were created with 30%, 60%, and 90% fewer mismatches.

Ten different example sets were created for each level of closeness to the domain theory: 10%, 30%, 60%, 90% fewer matches, and 30%, 60%, 90% fewer mismatches. In total, forty example sets were created which matched the original theory *less* closely than the original example set, and thirty example sets were created which matched the original theory *more* closely than the original example set. Each of these example sets was tested using a leave-one-out methodology using C4.5 and the TGCI algorithm. The results are summarized in Figure 21. The x-axis is a measure of *theory proximity* – closeness of an example set to the domain theory. "0" on the x-axis indicates no change in the original promoter examples. "100" on the x-axis means that each positive example exactly matches the domain theory. "-100" on the x-axis means that any match of a feature value of a positive example to the domain theory is totally by chance[4]. Each datapoint in Figure 21 is the result of averaging the accuracies of the ten example sets for each level of theory proximity (except for the point at zero which is the accuracy of the exact original promoter examples).

---

3. More precisely, there would be slightly more matches than 30% fewer matches because some features would be randomly reassigned back to their original matching value.

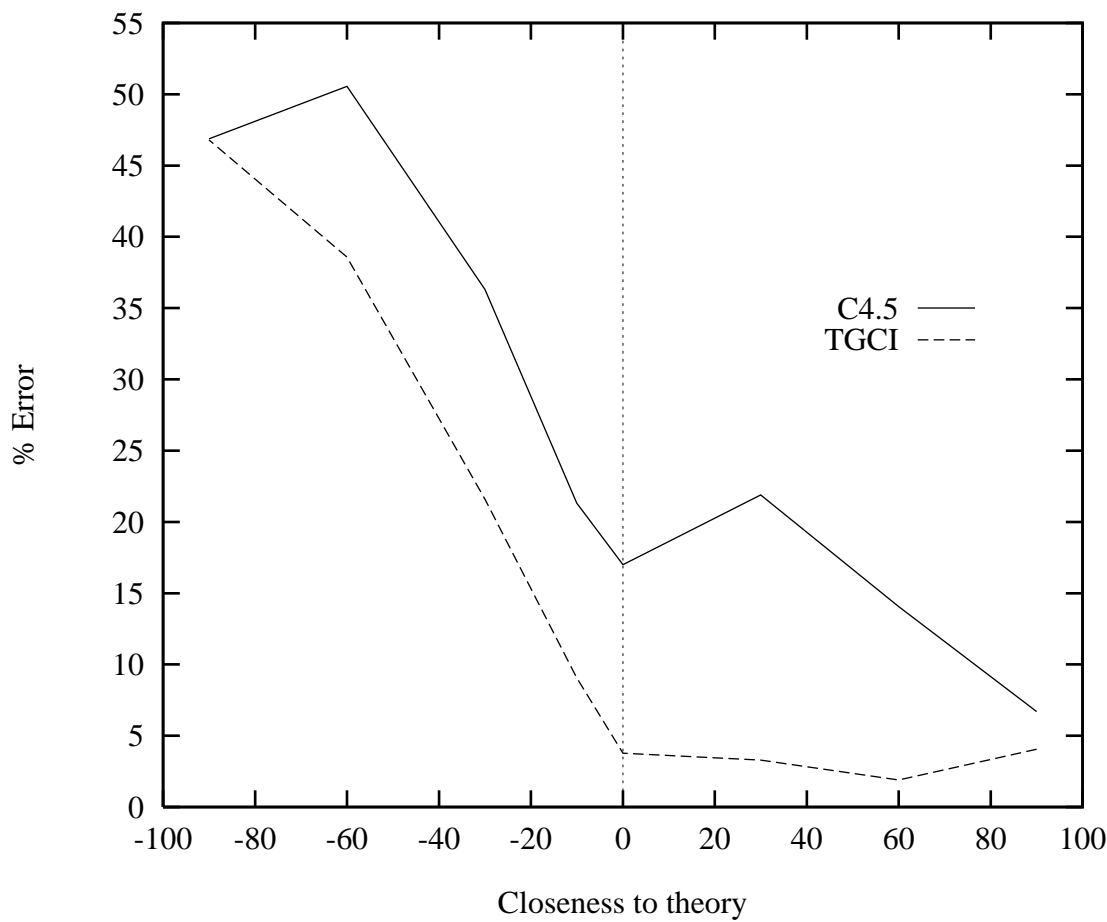

Figure 21: Seven altered promoter domains were created, three that more closely matched the theory than the original domains and four that less closely matched. A 100 on the x-axis indicates a domain in which the positive examples match the domain theory 100%. A negative 100 indicates a domain in which any match of the positive examples to the domain theory is purely chance. The accuracy of C4.5 and TGCI are plotted for different levels of proximity to the domain theory.

One notable portion of Figure 21 is the section between 0 and 60 on the x-axis. Domains in this region have a greater than trivial level of mismatch with the domain theory but not more than moderate mismatch. This is the region of TGCI's best performance. On these domains, TGCI achieves high accuracy while a standard learner, C4.5, using the original feature set gives mediocre performance. A second region to examine is between -60 and 0 on the x-axis where the level of mismatch ranges from moderate to extreme. In this region TGCI's performance falls off but its improvement over the original feature set remains high as shown in Figure 22 which plots the improvement of TGCI over C4.5. The final two regions to notice are greater than 60 and less than -60 on the x-axis. As the level of mismatch between theory and examples becomes trivially small (x-axis greater than 60), C4.5 is able to pick out the theory's patterns leading to high accuracy that approaches that of TGCI's. As the level of mismatch becomes extreme (x-axis less than -60) the theory gives little help in problem-solving resulting in similarly poor accuracy for both methods. In summary, as shown in Figure 22 for variants of the promoter problem there is a wide range of theory proximity (centered around the real promoter problem) for which theory-guided constructive induction yields sizable improvement over standard learners.

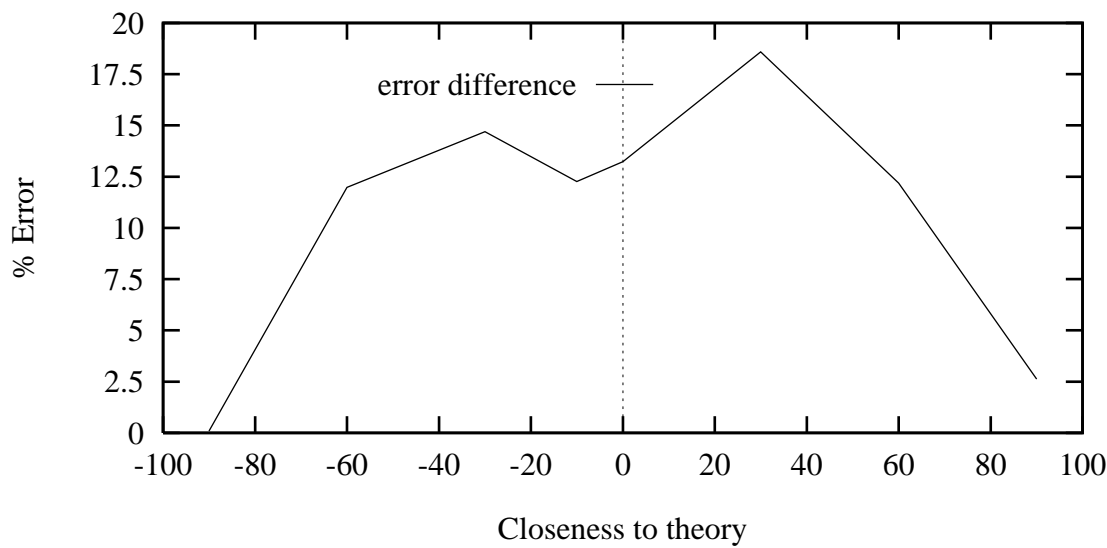

Figure 22: The difference in error between C4.5 and TGCI for different levels of proximity of the example set to the domain theory.

---

4. The scale 0 to -100 on the left half of the graph may not be directly comparable with the scale 0 to 100 on the right half of the graph since there were not a equal number of matches and mismatches in the original examples.

## 8. Conclusion

Our goal in this paper has not been just to present another new system, but rather to study the two qualities *flexible representation* and *flexible structure*. These capabilities are intended as a frame of reference for analyzing theory-guided systems. These two principles provide guidelines for purposeful design. Once we had distilled the essence of systems such as Miro, Kbann, and Neither-MofN, theory-guided constructive induction was a natural synthesis of their strengths. Our experiments have demonstrated that even a simple application of the two principles can effectively integrate theory knowledge with training examples. Yet there is much room for improvement; the two principles could be quantified and made more precise, and the implementations that proceed from them should be explored and refined.

Quantifying representational flexibility is one step. Section 4 gave three degrees of flexibility: one measured the exact match to a theory, one counted the number of matching conditions, and one allowed for a weighted sum of the matching conditions. The amount of flexibility should be quantified, and finer-grained degrees of flexibility should be explored. The accuracy in assorted domains should be evaluated as a function of representational flexibility.

Finer-grained structural flexibility would be advantageous. We have presented systems that make small, incremental modifications in a theory as lacking structural flexibility. Yet theory-guided constructive induction falls at the other extreme, perhaps allowing excessive structural flexibility. Fortunately, existing induction tools are capable of fashioning simple yet highly predictive theory structures when the problem features are suitably high-level. Nevertheless, approaches should be explored that take advantage of the structure of the initial theory without being unduly restricted by it.

The strength discussed in Section 6.5 should be given further attention. Although the promoter domain gives a very small example of synthesizing competing theories, this should be explored in a domain in which entire competing, inconsistent theories are available such as synthesizing the knowledge given by multiple experts. The point was made in Section 6.4 that Tgci can use theory fragments to evaluate the contribution of different parts of a theory. This should also be explored further.

In an exploration of bias in standard induction, Utgoff (1986) refers to biases as ranging from *weak* to *strong* and from *incorrect* to *correct*. A strong bias restricts the concepts that can be represented more than a weak bias thus providing more guidance in learning. But as a bias becomes stronger, it may also become incorrect by ruling out useful concept descriptions. A similar situation arises in theory revision — a theory representation language that is inappropriately rigid may impose a strong, incorrect bias on revision. A language that allows adaptability along too many dimensions may provide too weak a bias. A Grendel-like toolbox would allow a theory to be translated into a range of representations with varying dimensions of adaptability. Utgoff advocates starting with a strong, possibly incorrect bias and shifting to an appropriately weak and correct bias. Similarly, a theory could be translated into successively more adaptable representations until an appropriate bias is found. We have implemented only a single tool; many open problems remain along this line of research.

The converse relationship of theory revision and constructive induction warrants further examination — theory revision uses data to improve a theory; constructive induction can use theory to improve data to facilitate learning. Since the long-term goal of machine learning is to use data, inference, and theory to improve any and all of them, we believe that a consideration of these related methods can be beneficial, particularly because each research area has some strengths that the other lacks.

An analysis of landmark theory revision and theory-guided learning systems has led to the two principles *flexible representation* and *flexible structure*. Because theory-guided constructive induction was based upon these high-level principles, it is simple yet achieves good accuracy. These principles provide guidelines for future work, yet as discussed above, the principles themselves are imprecise and call for further exploration.

### Acknowledgements

We would like to thank Geoff Towell, Kevin Thompson, Ray Mooney, and Jeff Mahoney for their assistance in getting the datapoints for KBANN, LABYRINTH$_K$, and EITHER. We would also like to thank Paul Baffes for making the NEITHER program available and for advice on setting the program's parameters. We thank the anonymous reviewers for their constructive criticism of an earlier draft of this paper. We gratefully acknowledge the support of this work by a DoD Graduate Fellowship and NSF grant IRI-92-04473.